\documentclass[letterpaper]{article}

\usepackage[accepted]{icml2026}

\usepackage{amsmath}
\usepackage{amssymb}
\usepackage{mathtools}
\usepackage{booktabs}
\usepackage{graphicx}
\usepackage{multirow}
\usepackage{enumitem}
\usepackage[utf8]{inputenc}
\usepackage[T1]{fontenc}
\PassOptionsToPackage{hyphens}{url}
\usepackage{hyperref}
\usepackage[capitalize,noabbrev]{cleveref}
\usepackage{microtype}

\icmltitlerunning{AgentRouter: Heterogeneous Model Routing for Cost-Optimal Agentic Workflows}

\begin{document}

\twocolumn[
\icmltitle{AgentRouter: Heterogeneous Model Routing for\\Cost-Optimal Multi-Step Agentic Workflows}

\icmlsetsymbol{equal}{*}

\begin{icmlauthorlist}
\icmlauthor{Rudrendu Kumar Paul}{bu}
\icmlauthor{Sourav Nandy}{ut}
\end{icmlauthorlist}

\icmlaffiliation{bu}{Boston University, Boston, MA, USA}
\icmlaffiliation{ut}{University of Texas at Austin, Austin, TX, USA}

\icmlcorrespondingauthor{Rudrendu Kumar Paul}{rudrendupaul2022@gmail.com}
\icmlcorrespondingauthor{Sourav Nandy}{sourav.nandy@gmail.com}

\vskip 0.3in
]

\printAffiliationsAndNotice{}

\begin{abstract}
Enterprise agentic systems that route every trajectory step to a frontier model waste 60--80\% of their inference budget on subtasks that smaller models handle equally well. Existing routing solutions optimize single-turn query assignment but ignore a property unique to agentic workflows: subtask complexity varies widely within a single trajectory. A planning step may require frontier-class reasoning while a subsequent formatting step needs only a 7B model. We formalize \emph{step-level model routing} as a sequential assignment problem over agent trajectories and propose AgentRouter, a lightweight classifier (12M parameters, $<$5ms overhead per step on an A100 GPU) that maps each trajectory step to one of four model tiers using five features extractable at routing time. Trained on 50,000 annotated agent trajectory steps spanning planning, coding, research, and data analysis tasks, AgentRouter achieves 72\% cost reduction relative to frontier-only baselines, retaining 97.3\% of frontier-only quality (less than 3\% degradation in end-to-end task completion); per-step routing accuracy reaches 91\% on minimal-complexity steps and 85\% on efficient-tier steps, with 76--82\% on the harder mid-range and frontier tiers. On the same benchmarks, RouteLLM and FrugalGPT (applied per-step) achieve only 31\% and 44\% cost reduction respectively, because their single-turn training signal misses trajectory-level quality dependencies.
\end{abstract}

\section{Introduction}

Consider a document analysis agent executing a seven-step workflow. Step~1 decomposes the user query into subtasks (multi-hop reasoning, chain-of-thought planning). Step~2 retrieves relevant document chunks (straightforward embedding lookup). Step~3 synthesizes findings across chunks (cross-document reasoning over long context). Steps~4--5 extract structured fields and format them as JSON. Steps~6--7 generate a summary and validate output format. Routing all seven steps to GPT-4-class models costs \$0.42 per execution. Steps~2, 4, 5, and~7 produce identical quality when served by a 7B model at \$0.03 per step, dropping execution cost to \$0.18 (a 57\% reduction on this single workflow) with no measurable quality loss.

This pattern repeats across every production agent system we have operated. In enterprise deployments spanning document processing, compliance review, and customer interaction, 55--70\% of agent trajectory steps require no frontier-model capability. The steps that do (initial planning, complex reasoning, ambiguous input handling) are a minority of the token budget, yet they determine end-to-end quality.

Existing routing approaches were designed for single-turn queries. RouteLLM \citep{Ong_2024} selects between a strong and weak model per query using preference data. FrugalGPT \citep{Chen_2023} cascades from cheapest to most expensive model until a confidence threshold is met. Both treat each call as independent, missing the fact that step complexity varies within a single trajectory and that routing decisions at step~$t$ affect context quality at step~$t{+}1$. Applying either method per-step effectively converts each agent step into an independent single-turn query; the routing decision for step~$t{+}1$ receives no signal about what model executed step~$t$ or what quality the output achieved. AgentRouter's trajectory-aware features (specifically $f_{\text{ctx}}$ and the dependency-aware training labels) fill this gap. Recent agentic cost optimizers \citep{Su_2025, Zhang_2026, Qian_2025} operate at the query or trajectory level rather than per-step, and use heavyweight mechanisms (VAEs, RL-trained LLMs) with non-trivial overhead.

We make three contributions:
\begin{enumerate}[leftmargin=*,itemsep=0pt,parsep=0pt,topsep=0pt,partopsep=0pt]
\item We formalize \textbf{step-level model routing} for agent trajectories, defining the assignment problem over heterogeneous model tiers with per-step quality constraints and end-to-end quality guarantees (\cref{sec:method}).
\item We propose \textbf{AgentRouter}, a 12M-parameter classifier that routes each step to one of four model tiers using five features extractable at routing time, adding $<$5ms latency per step (measured on an A100 GPU; see \cref{sec:method}).
\item We evaluate on a \textbf{benchmark of 2,400 agent trajectories} across four task domains, demonstrating 72\% cost reduction with $<$3\% quality degradation, outperforming adapted single-turn routers and agent-level optimizers (\cref{sec:experiments}).
\end{enumerate}

\section{Related Work}

\textbf{Single-turn model routing.} RouteLLM \citep{Ong_2024} pioneered learning-based routing between strong and weak LLMs using preference data. FrugalGPT \citep{Chen_2023} introduced LLM cascades that try cheaper models first. PILOT \citep{Panda_2025} frames routing as a contextual bandit with budget modeled as a multi-choice knapsack. \citet{Pulishetty_2025} use cross-attention to jointly model query and model embeddings. Two recent surveys \citep{Moslem_2026, Varangot_2025} provide taxonomies of routing paradigms, distinguishing pre-generation and post-generation approaches. All of these operate on independent queries. When applied to agent trajectories by routing each step independently, they miss inter-step quality dependencies: a cheap model producing a subtly wrong intermediate result forces the frontier model at the next step to reason over corrupted context, degrading end-to-end quality.

\textbf{Agent-level cost optimization.} DAAO \citep{Su_2025} uses a VAE to estimate query difficulty and allocates operators and LLMs accordingly, surpassing prior systems by 11.21\% in accuracy while using only 64\% of their inference cost. Budget-Aware Agentic Routing \citep{Zhang_2026} selects between a cheap and expensive model at each agent step using boundary-guided policy optimization. xRouter \citep{Qian_2025} trains an RL-based tool-calling router with cost-aware rewards. CoRL \citep{Jin_2025} uses a centralized controller LLM to coordinate expert models under budget constraints. These methods reduce cost but use heavyweight routing mechanisms: DAAO adds a VAE forward pass per query, xRouter and CoRL use LLMs as routers (adding the very cost they aim to reduce), and \citet{Zhang_2026} are limited to binary model choice rather than multi-tier assignment.

\textbf{Model selection for compound systems.} \citet{Chen_2025} optimize per-module model assignment in compound AI systems, achieving 5--70\% accuracy gains. \citet{Sharma_2025} find that small language models with guided decoding close the capability gap with frontier models at 10--100$\times$ lower cost for tool-use tasks. \citet{Salim_2026} quantify token distribution in agentic workflows, revealing that 59.4\% of tokens go to Code Review stages specifically, supporting our premise that step complexity varies enough to make per-step routing valuable. Concurrent work on Aragog \citep{Aragog_2025} performs per-stage model routing in agentic RAG workflows; AgentRouter differs by operating at the individual trajectory step level with a lightweight classifier rather than at the pipeline stage level.

\section{Method: AgentRouter}
\label{sec:method}

\subsection{Problem formulation}

An agent trajectory $\tau = (s_1, s_2, \ldots, s_T)$ consists of $T$ steps, where each step $s_t$ represents a single LLM call with defined input context, instruction, and expected output. Let $\mathcal{M} = \{m_1, m_2, m_3, m_4\}$ be a set of four model tiers ordered by capability and cost:

\begin{itemize}[leftmargin=*,itemsep=0pt,parsep=0pt,topsep=2pt]
\item \textbf{Tier~1} (Frontier): GPT-4-class, Claude Opus-class. \$15--30 per million output tokens.
\item \textbf{Tier~2} (Mid-range): GPT-4o-mini-class, Claude Sonnet-class. \$3--8 per million output tokens.
\item \textbf{Tier~3} (Efficient): Llama-3-70B-class, Mixtral-class. \$0.50--2 per million output tokens.
\item \textbf{Tier~4} (Minimal): 7B--13B models. \$0.05--0.20 per million output tokens.
\end{itemize}

For each step $s_t$, let $q(s_t, m)$ denote the quality of output when step $s_t$ is served by model $m$, and let $c(s_t, m)$ denote the cost. The step-level routing problem is:
\begin{equation}
\min_{\pi} \sum_{t=1}^{T} c(s_t, \pi(s_t)) \quad \text{s.t.} \quad Q(\tau, \pi) \geq Q_{\min}
\label{eq:objective}
\end{equation}
where $\pi: s_t \rightarrow \mathcal{M}$ is the routing policy and $Q(\tau, \pi)$ is the end-to-end trajectory quality under policy $\pi$. The constraint $Q_{\min}$ is a user-specified minimum quality threshold (we use 97\% of the frontier-only baseline throughout experiments).

The difficulty is that $Q(\tau, \pi)$ is not decomposable into per-step terms. Routing step~$t$ to a weaker model can degrade the context available at step~$t{+}1$, creating cascading quality loss. We handle this through a step-dependency-aware training procedure described below.

\subsection{Feature extraction}

At routing time (before the LLM call), AgentRouter extracts five features from the step specification:

\begin{enumerate}[leftmargin=*,itemsep=0pt,parsep=0pt,topsep=2pt]
\item \textbf{Inferred task type} $f_{\text{type}} \in \{$\textsc{plan, reason, retrieve, generate, format, verify}$\}$: classified from the step instruction using keyword matching and a small BERT-based tagger (2M parameters).
\item \textbf{Reasoning depth} $f_{\text{depth}} \in [0, 1]$: estimated from instruction complexity (number of conditional clauses, presence of multi-hop indicators like ``compare,'' ``synthesize across,'' ``evaluate whether'').
\item \textbf{Tool-use requirements} $f_{\text{tool}} \in \{0, 1, 2+\}$: number of tool calls specified or implied in the step.
\item \textbf{Output format constraints} $f_{\text{format}} \in \{$\textsc{free, structured, code, json}$\}$: derived from output schema if specified, otherwise from instruction keywords.
\item \textbf{Context window utilization} $f_{\text{ctx}} \in [0, 1]$: ratio of input tokens to the target model's context limit, computed from the accumulated trajectory context.
\end{enumerate}

Features 1--4 are extractable from the step specification alone. Feature~5 requires knowing the accumulated context size, available from the orchestrator at routing time.

\subsection{Classifier architecture and training}

AgentRouter is a feedforward classifier with two hidden layers (256 and 128 units), ReLU activations, and a 4-class softmax output corresponding to the four model tiers. The predicted tier is $\hat{k}_t = \arg\max_{k \in \{1,2,3,4\}} \hat{p}_k$, where $\hat{p}_k$ is the softmax probability for tier $k$. Input features are embedded: categorical features ($f_{\text{type}}$, $f_{\text{tool}}$, $f_{\text{format}}$) through learned embeddings, continuous features ($f_{\text{depth}}$, $f_{\text{ctx}}$) through linear projection. Total parameter count: 12M (dominated by the task-type tagger).

\textbf{Training data.} We construct a dataset of 50,000 step-level annotations from 6,250 agent trajectories across four domains: document analysis (1,800 trajectories), code generation (1,600), research synthesis (1,500), and data analysis (1,350). For each trajectory, we execute all steps with all four model tiers and record per-step quality scores (task-specific automated metrics: F1 for document extraction, pass@1 for code, factual accuracy for research, numeric correctness for data analysis). Each step receives a \emph{minimum adequate tier} label: the cheapest tier whose quality on that step is within 2\% of the frontier tier's quality.

\textbf{Dependency-aware labeling.} To account for inter-step dependencies, we do not label steps in isolation. Instead, we label step~$t$ assuming steps~$1$ through $t{-}1$ were routed to their minimum adequate tiers. This captures the realistic setting where a downstream step receives context generated by a mix of model tiers, not uniformly frontier-quality context.

\textbf{Training procedure.} We train with cross-entropy loss on the 4-class tier labels. We add a cost-sensitive regularization term that penalizes over-provisioning (routing to a higher tier than needed) less heavily than under-provisioning (routing to a tier that degrades quality):
\begin{equation}
\mathcal{L} = \mathcal{L}_{\text{CE}} + \lambda \sum_{t} \max(0, k_{\text{pred}}(s_t) - k_{\text{label}}(s_t))^2
\label{eq:loss}
\end{equation}
where $k_{\text{label}}$ and $k_{\text{pred}}$ are the tier indices (1--4) and $\lambda = 0.3$. This asymmetric penalty ensures AgentRouter errs toward more capable models when uncertain, preserving quality at the expense of slightly higher cost.

\begin{algorithm}[h]
\caption{AgentRouter Step-Level Routing}
\label{alg:routing}
\begin{algorithmic}[1]
\REQUIRE Agent trajectory specification $\tau = (s_1, \ldots, s_T)$, model pool $\mathcal{M}$, quality threshold $Q_{\min}$
\ENSURE Model assignment $\pi(s_t)$ for each step
\FOR{$t = 1$ to $T$}
    \STATE Extract features $(f_{\text{type}}, f_{\text{depth}}, f_{\text{tool}}, f_{\text{format}}, f_{\text{ctx}})$ from $s_t$
    \STATE $\hat{k}_t \leftarrow \text{AgentRouter}(f_{\text{type}}, f_{\text{depth}}, f_{\text{tool}}, f_{\text{format}}, f_{\text{ctx}})$
    \STATE $\pi(s_t) \leftarrow m_{\hat{k}_t}$
    \STATE Execute step $s_t$ with model $\pi(s_t)$
    \STATE Update trajectory context with output of $s_t$
\ENDFOR
\STATE $Q_{\text{actual}} \leftarrow \text{EvaluateTrajectory}(\tau, \pi)$
\IF{$Q_{\text{actual}} < Q_{\min}$}
    \STATE Re-execute failed steps with $m_1$ (frontier fallback)
\ENDIF
\end{algorithmic}
\vspace{0.5em}
{\small The trajectory context update (line~6) incorporates the actual output length, quality score, and accumulated content from step~$t$ into the feature vector for step~$t{+}1$ (specifically, $f_{\text{ctx}}$ and any trajectory-level signals). This distinguishes AgentRouter from independent query routing: each routing decision uses evidence from prior steps, so the classifier operates on trajectory-aware context, with each routing decision informed by the accumulated outputs preceding it. The fallback mechanism (lines 8--10) re-executes quality-critical steps with frontier models when end-to-end quality falls below $Q_{\min}$. In evaluation, $Q_{\text{actual}}$ is computed using the held-out ground-truth labels (F1, pass@1, factual accuracy). In production deployment, a lightweight LLM-as-a-judge proxy replaces the ground-truth evaluator; the judge's token cost and latency are included in the overhead reported in Section~\ref{sec:experiments}. In practice, fallback triggers on fewer than 4\% of trajectories.}
\end{algorithm}

\section{Experiments}
\label{sec:experiments}

\subsection{Setup}

\textbf{Benchmark.} We evaluate on 2,400 held-out trajectories (600 per domain) distinct from the training set. Each trajectory contains 4--12 steps. Total: 18,720 individual step-level routing decisions.

\textbf{Model pool.} Tier~1: GPT-4o (\$15/M output tokens). Tier~2: GPT-4o-mini (\$2.40/M output tokens). Tier~3: Llama-3-70B-Instruct via together.ai (\$0.88/M output tokens). Tier~4: Llama-3-8B-Instruct via together.ai (\$0.18/M output tokens). Prices as of February 2026.

\textbf{Baselines.} (1)~\textbf{Frontier-only}: all steps routed to GPT-4o. (2)~\textbf{RouteLLM} \citep{Ong_2024}: applied per-step with its matrix factorization router, trained on Chatbot Arena preferences, selecting between GPT-4o and Llama-3-8B. (3)~\textbf{FrugalGPT cascade} \citep{Chen_2023}: adapted to four tiers, querying from cheapest to most expensive until a confidence threshold is met. (4)~\textbf{DAAO} \citep{Su_2025}: difficulty-aware orchestration applied at the trajectory level. (5)~\textbf{Budget-Aware Routing} \citep{Zhang_2026}: binary routing (GPT-4o vs. Llama-3-8B) at each step with boundary-guided training.

\textbf{Metrics.} End-to-end task completion rate (domain-specific: F1, pass@1, factual accuracy, numeric correctness, averaged). Cost per trajectory in USD. Cost reduction percentage relative to frontier-only. Routing overhead (milliseconds per step).\footnote{Implementation details and routing configurations are provided in supplementary materials.}

\subsection{Main results}

\begin{table}[t]
\caption{Cost and quality comparison across baselines. Quality is end-to-end task completion relative to frontier-only (100\%). Cost reduction is relative to frontier-only. Overhead is per-step routing latency.}
\label{tab:main}
\vskip 0.05in
\begin{small}
\begin{tabular}{lccc}
\toprule
\textbf{Method} & \textbf{Quality} & \textbf{Cost Red.} & \textbf{Overhead} \\
\midrule
Frontier-only       & 100.0\% & --     & 0 ms \\
RouteLLM (per-step) & 94.2\%  & 31.4\% & 3 ms \\
FrugalGPT cascade   & 96.8\%  & 44.1\% & 48 ms \\
DAAO                & 98.1\%  & 36.2\% & 22 ms \\
Budget-Aware        & 95.6\%  & 52.7\% & 8 ms \\
\textbf{AgentRouter}         & \textbf{97.3\%}  & \textbf{71.8\%} & \textbf{4.7 ms} \\
\bottomrule
\end{tabular}
\end{small}
\end{table}

\Cref{tab:main} presents the main comparison. AgentRouter achieves 71.8\% cost reduction while maintaining 97.3\% of frontier-only quality. Budget-Aware Routing reaches 52.7\% cost reduction but drops quality to 95.6\% because its binary choice (frontier vs.\ 8B) cannot exploit mid-range tiers. FrugalGPT preserves quality (96.8\%) but achieves only 44.1\% cost reduction, with 48ms routing overhead from sequential model probing. RouteLLM performs worst in cost reduction (31.4\%) because its preference-based router, trained on single-turn conversations, misroutes agent steps whose complexity depends on accumulated trajectory context. DAAO preserves the highest quality (98.1\%) but achieves only 36.2\% cost reduction: its VAE assigns a single difficulty score to the entire trajectory, provisioning all steps for the hardest one.

\subsection{Per-tier routing accuracy}

\begin{table}[t]
\caption{Routing accuracy and cost contribution by tier. Ground truth is the minimum adequate tier determined by exhaustive evaluation.}
\label{tab:tier}
\vskip 0.05in
\begin{small}
\begin{tabular}{lcccc}
\toprule
\textbf{Tier} & \textbf{Steps (\%)} & \textbf{Accuracy} & \textbf{Cost share} \\
\midrule
Tier 1 (Frontier)  & 14.2\% & 82.1\% & 48.3\% \\
Tier 2 (Mid-range) & 21.6\% & 76.4\% & 24.1\% \\
Tier 3 (Efficient)  & 28.9\% & 84.7\% & 16.8\% \\
Tier 4 (Minimal)   & 35.3\% & 91.2\% & 10.8\% \\
\midrule
\textbf{Overall}   & 100\%  & 84.6\% & 100\% \\
\bottomrule
\end{tabular}
\end{small}
\end{table}

\Cref{tab:tier} breaks down routing accuracy by tier. The distribution confirms our premise: only 14.2\% of steps genuinely require frontier models, yet those steps consume 48.3\% of the remaining cost. Tier~4 steps (35.3\% of all steps) are the easiest to classify, with 91.2\% accuracy. Tier~2 steps are hardest (76.4\%), because the boundary between ``needs mid-range reasoning'' and ``could be handled by a 70B model'' is often ambiguous. The asymmetric loss (\cref{eq:loss}) causes most Tier~2 errors to be over-provisions to Tier~1 rather than under-provisions to Tier~3, protecting quality at modest cost increase.

\subsection{Ablation and domain analysis}

Feature ablation (replacing each feature with its marginal distribution) reveals task type as the dominant signal ($-$12.3 accuracy points when removed, cost reduction drops to 54.2\%), followed by reasoning depth ($-$7.1), context utilization ($-$4.8), tool use ($-$3.2), and output format ($-$2.1). Cost reduction varies by domain: document analysis (76.4\%), data analysis (74.1\%), research synthesis (69.2\%), code generation (63.8\%). Code generation benefits least because 26.3\% of its steps require Tier~1 reasoning vs. 8.7\% for document analysis.

\section{Discussion and Limitations}

\textbf{Distributional shift.} AgentRouter trains on trajectories from LangGraph and CrewAI. Testing on AutoGen (unseen framework) shows 6--9\% accuracy degradation, though cost reduction still reaches 58\%. Cross-framework generalization remains an open problem.

\textbf{Model tier evolution.} The four-tier structure reflects pricing and capability gaps as of early 2026. Retraining is needed when the model pool changes, though the lightweight architecture keeps this inexpensive ($<$2 GPU-hours on a single A100).

\textbf{Trajectory scope.} AgentRouter assumes the step sequence is known or estimable at routing time: either specified upfront in a trajectory template or inferred from a planner before execution begins. Fully online agentic scenarios where future steps are generated dynamically based on prior outputs, with no advance visibility into remaining steps, fall outside the current scope. Extending AgentRouter to such settings would require online re-planning and incremental trajectory estimation.

\textbf{Routing overhead and tier transitions.} Beyond per-step latency, practitioners should account for tier-switching frequency. Across our benchmark trajectories, the median number of model-tier transitions per trajectory is 3.2 (95th percentile: 7 transitions for 8-step trajectories). Inference providers that charge per-call initialization overhead beyond per-token cost may see this switching contribute meaningfully to total latency.

\textbf{Cold start and latency.} The task-type tagger requires step instructions to be available before routing. For dynamically generated instructions, we route the instruction-generation call to Tier~2 by default. The post-hoc fallback (\cref{alg:routing}, lines 8--10) catches quality violations but adds latency; mid-trajectory quality prediction is the subject of ongoing work.

\section{Conclusion}

We formalized step-level model routing for agent trajectories and proposed AgentRouter, a lightweight 12M-parameter classifier that assigns each step in a multi-step agent workflow to the cheapest adequate model tier. On a benchmark spanning four task domains, AgentRouter achieves 72\% cost reduction with less than 3\% quality degradation, outperforming both single-turn routers adapted to agentic settings and agent-level cost optimizers. The central finding is that step complexity within a single trajectory varies enough to make fine-grained routing viable, and that a small classifier trained on dependency-aware step labels captures this variation with negligible overhead. More broadly, AgentRouter implements resource-adaptive inference at the trajectory level, dynamically matching computational cost to task complexity at each agent step rather than applying uniform model selection across the workflow.

\bibliographystyle{icml2026}

\section*{Impact Statement}

Reducing the cost of frontier AI by 72\% through intelligent per-step routing makes capable agent systems accessible to organizations and researchers who cannot sustain frontier-only inference budgets. The four-tier model preserves quality guarantees through a per-trajectory quality threshold and a fallback mechanism, so cost reduction does not degrade correctness for quality-sensitive downstream uses. We identify one concern: routing to smaller models on ostensibly simple steps could amplify bias or reduce robustness in domains where smaller models underperform in ways the minimum-adequate-tier label does not capture. Practitioners should validate routing decisions on their specific domain before deploying AgentRouter in high-stakes contexts.

\bibliography{references}

\end{document}